# Global-Local Medical SAM Adaptor Based on Full Adaption

Meng Wang
School of Electronic and Information Engineering
Liaoning Technical University
Xingcheng City, Liaoning Province, P. R. China
1163855905@qq. com

Yarong Feng
School of Electronic and Information Engineering
Liaoning Technical University
Xingcheng City, Liaoning Province, P. R. China
18093490172@163. com

Yongwei Tang
School of Electronic and Information Engineering
Liaoning Technical University
Xingcheng City, Liaoning Province, P. R. China
2458672708@qq. com

Yuxin Liang
School of Electronic and Information Engineering
Liaoning Technical University
Xingcheng City, Liaoning Province, P. R. China
2833391387@qq. com

Tian Zhang
Software college
Northeastern University
Shenyang, Liaoning Province, P. R. China
zhangtnnn@163. com

Chao Lv*
Department of General Surgery, Shengjing Hospital
China Medical University
Shenyang, Liaoning Province, P. R. China
clu@cmu. edu. cn

**Abstract**一 Visual language models (VLM) have made great breakthroughs in universal semantic segmentation and significantly aid the improvements of many domain-specific down-stream tasks. Particularly, with the help of parameter-efficient adaption techniques, such as Medical SAM adaptor (Med-SA), VLM can be well adapted to medical image segmentation. However, Med-SA still can be improved, as it fine-tunes VLM in a partial adaption manner where the adapters are separated by frozen layers and lack of collaboration in back-propagation. To resolve this problem, we present a full adaption technique and a novel global medical SAM adaptor (GMed-SA) which can adapt SAM globally. Specifically, all adapters are connected by full adaption, enabling them to work collaboratively for VLM adaption. We further combine GMed-SA and Med-SA to propose a global-local medical SAM adaptor (GL-Med-SA). The combined method adapts SAM both globally and locally, and slightly improve the segmentation performance. We performed extensive experiments based on a challenging publicly available 2D melanoma dataset, and the experimental results showed that GL-Med-SA outperforms several state-of-the-art methods in both Dice metrics and Intersection over union (IoU), demonstrating our methods` superiority.

*Segment anything model; medical SAM adaptor; melanoma; semantic segmentation; adaption*

I. INTRODUCTION

The developments of artificial intelligence (AI) have significantly aided the field of medical image analysis, especially the presentation of various visual language models. They have been instrumental in minimizing the difference between vision and language tasks, enabling machines to understand multiple types of prompts and generate natural language descriptions related to visual content. Among these models, the Segment Anything Model (SAM) [1] is reported as a powerful visual language model that stands out for its remarkable capabilities in universal object segmentation. Inspired by SAM, extensive methods have promoted numerous applications in various domains. In particular, they have revolutionized the field of image segmentation and outperform many state-of-the-art (SOTA) semantic segmentation methods in applications with comprehend images. This is mainly due to the combination of various types of neural networks, (e.g., UNet and CLIP), SAM`s power of zero-shot generalization ability, and Parameter-Efficient Fine-Tuning methods (e.g., Adaption).

As one of the most successful SAM extensions, Medical SAM adaptor (Med-SA) [4], an advanced segmentation model based on SAM and adaption techniques, has been developed for universal medical image segmentation. This model leverages zero-shot generalization ability of SAM in semantic segmentation and is specifically optimized for both 2D and 3D medical image segmentation. However, despite its remarkable capabilities, the Medical SAM adaptor may have some limitations. One of the possible drawbacks is that each adaptor module independently adjusts the network`s continuous layers, instead of jointly adjusting the entire network of SAM. This limitation may make it difficult to focus on some specific layers that require more global attention to adapt, and it restricts the Med-SA`s ability to fit for specific tasks, leading to room for further improvements in efficiency and performance.

In this paper, to further improve Medical SAM adaptor for accurate and efficient adaptation of SAM, we introduce a novel Global Medical SAM Adaptor (GMed-SA), that globally adapts SAM with a hierarchical neural network. To further enhance the performance, we combine GMed-SA with Med-SA to propose a Global-Local Medical SAM Adaptor (GL-Med-SA), which can adapt SAM both globally and locally. To demonstrate the effectiveness of GL-Med-SA, we conduct extensive experiments on a challenging public 2D melanoma segmentation dataset, and we show the results in the rest of the paper both

quantitatively and qualitatively. In summary, the key contributions in this paper include:

The proposal of a novel Global Medical SAM Adaptor (GMed-SA) that globally adjusts SAM's parameters to better accommodate the features of various medical images and the rigorous requirements of semantic segmentation.

The design of a Global-Local Medical SAM Adaptor (GL-Med-SA), that combines the advantages of both GMed-SA and Med-SA, adapting SAM globally and locally simultaneously to further improve the performance.

We organize the remaining of this paper as follows. Section II describes the main ideas and core techniques of related Interactive segmentation models and PEFT, as well as their advantages and disadvantages. Section III details the insights and contents of our methods. Section IV describes the experimental details, shows the results, and makes preliminary analysis based on them. Section V makes deeper analysis and discussions. Finally, Section VI concludes the paper and gives possible future directions.

## II. RELATED WORKS

*Interactive segmentation methods*:

Interactive segmentation has become an influential method in the realm of visual comprehension. This kind of methods has seen a surge in interest across multiple fields, including computer vision and medical image processing. They empower models to produce a wide range of precise segmentation masks in response to user inputs. Pioneering studies such as DIOS[5] showcased the promise of interactive segmentation for object delineation. However, it is the more contemporary approaches like SAM[1], that have significantly propelled the field forward by integrating advanced techniques such as prompt engineering, confidence scoring, and multi-pointer interactions, resulting in more reliable and precise segmentation.

The advent of SAM introduced a groundbreaking concept known as promptable segmentation. In this novel paradigm, the model is designed to generate valid segmentation masks in response to any form of segmentation prompt—be it points, bounding boxes, masks, or even free-form text. This innovation opens up avenues for zero-shot, one-shot, and few-shot learning across unseen tasks or datasets, positioning promptable segmentation as a frontier for foundational model development in computer vision.

Building on the success of SAM, subsequent advancements like Efficient SAM and Mobile SAM [6] have emerged, focusing on optimizing the original SAM architecture for efficiency and mobile deployment, respectively. These extensions aim to broaden the applicability of SAM by reducing computational demands and enabling real-time segmentation on resource-constrained devices. Efficient SAM[7] streamlines the model's architecture to enhance speed without compromising much on accuracy, while Mobile SAM[6] is specifically designed for mobile platforms, ensuring that the benefits of SAM's interactive segmentation capabilities are accessible even in portable and embedded systems. Together, these developments underscore the versatility and potential of SAM-based models in the evolving landscape of computer vision technologies. Considering that the characteristic of SAM is especially suitable for the computer-aided organ and lesion segmentation in the clinical practice, Med-SA apply a new adaption technology to utilize the effectiveness of SAM in managing ambiguity and offering detailed control over medical segmentation outcomes.

*Parameter-Efficient Fine-Tuning*:

PEFT techniques have emerged as an efficient learning strategy for LLM and VLM Adaption [3], and they have demonstrated significant success in natural language processing (NLP) and computer vision. These techniques allow for fine-tuning large pre-trained language models using only a small number of additional parameters, while they preserve the generalization capabilities of the based models. Inspired by PEFT, Med-SA further develops an efficient medical SAM adapter. As a result, it only updates about 2% of SAM's parameters, while achieving significant improvements over other medical image segmentation methods including full-parameter fine-tuning ones. Specifically, the adapters are composed of multi-layer perceptron (MLP) and placed at different positions in the neural network. Other key innovations includes a Space-Depth Transpose (SD-Trans) Adapter and a Hyper-Prompting Adapter (HyP-Adpt) [4]. SD-Trans adapts 2D SAM to 3D medical images, and it can effectively capture the spatial and depth correlations in 3D medical data, enabling more accurate segmentation of complex structures. HyP-Adpt facilitates prompt-conditioned adaptation, allowing the model to effectively incorporate domain-specific medical knowledge and further improve segmentation performance.

## III. METHODOLOGY

This section first looks back into SAM and Med-SA, then it describes the architecture of GMed-SA, and finally it explains how they are integrated to form GL-Med-SA.

### A. Preliminary

The main architecture of SAM contains three parts including an image encoder to extract global correlation across different patches of the whole inoput images, a prompt encoder to integrate context information from user input prompts of different modalities, and a light-weight transformer based mask decoder to generate segmentation masks. The image encoder is a MAE pre-trained standard vision transformer, which is configured with 14×14 window-sized attention layers and 4 equally-spaced transformer blocks [4]. In prompt encoder, encoders are specifically designed for point prompts, mask prompts, and texts based on convolutional modules, the off-the-shelf CLIP tokens and the sum of positional encoding and learnable embedding. The mask decoder employs a two-way image-prompt cross attention transformer decoding manner, followed by a dynamic MLP to output tokens for semantic predictions and IOU scores. SAM enables users to employ various types of prompts during inference, and it combines random and iterative click sampling strategies to simulate user interaction and improve the model's robustness and accuracy during training.

### B. Medical SAM adaptor

Instead of fully adjusting all parameters [10], Med-SA is designed to adapt SAM minimally. It utilizes PEFT techniques to update only a small fraction of SAM's parameters, resulting in robust, efficient, and effective adaptation for domain-specific medical knowledge.

As aforementioned, Med-SA freezes SAM's parameters, and it designs a trainable module termed as MLP Adapter, which is then delicately integrated into specified locations of SAM. This Adapter functions as a bottleneck architecture, comprising a sequence of operations: a fully connected layer projecting embedding features down, a ReLU activation function, and a fully connected layer projecting embedding features up.

Specifically, there are three distinct Adapter modules used within the SAM encoder and decoder, each with a similar structure, placement, and function. We describe them in detail to better review Med-SA.

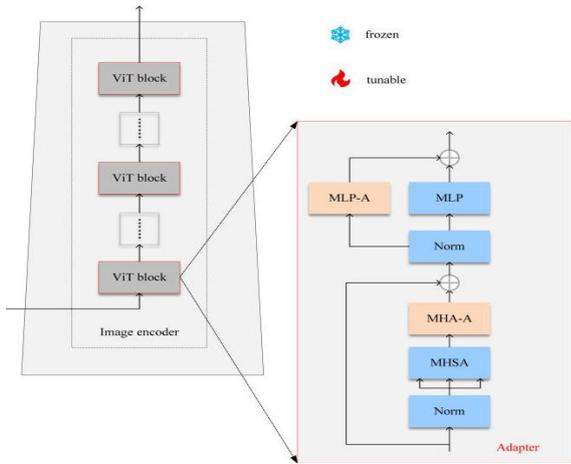

Figure 1. Schematic diagram of a Med-SA adapter.

Enhancement Adapter (MLP-A), as depicted in figure 1, is placed right behind the multi-head attention layers of image encoder and in the residual path of the corresponding MLP layers, and it is placed in the same manner as it is in the decoder. MLP-A serves to adapt and fine-tune the embedding processed by the MLP layer, potentially acting as a bottleneck to improve parameter efficiency and adaptability.

Multi-Head Attention Adapter (MHA-A) is similar to the MLP Enhancement Adapter, except that in the encoder path, it is inserted right after MHA of ViT blocks. MHA-A modifies the output embedding of the multi-head attention mechanism, allowing for more flexible and efficient transformation of the attention outputs.

Hyper-Prompting Adapter is a novel structure specifically designed to incorporate prompt embedding, so as to condition the model's generation or processing based on the provided prompts and adapt the prompt embedding learning from source tasks to the down-stream tasks. To this end, HyP-Adpt uses MLP to generate weights multiplied (via matrix multiplication) to the prompt embedding adapted by MLP-A.

Key components of Med-SA also includes SD-Trans, which enables the adaptation of 2D SAM to 3D medical images by transposing the spatial dimension to the depth dimension of inputs to learn spatial and depth correlations simultaneously. Note that, in this paper, we only adopt the adaption in the image encoder.

### C. Global Medical SAM adaptor

Regardless of the performance, Med-SA, as depicted in figure 1 (b), adapts SAM locally. This is indicated by the architecture of aforementioned adapters, including MLP-A, MHA-A, and HyPAdpt, which are responsible for fine-tuning several sequential layers in the whole SAM architecture and lack of collaboration.

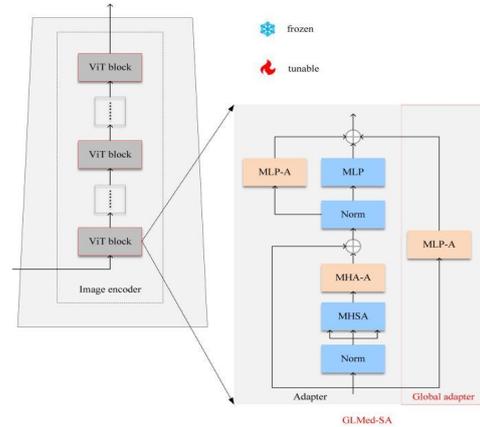

Figure 2. The structure of an adapter in GL-Med-SA. As seen, GL-Med-SA builds a high-way using hierarchically stacked adapters. Therefore it can perform full adaption across the whole network architecture.

To adapt SAM from a global perspective view, we propose Global Medical SAM Adaptor (i.e., GMed-SA), which is a kind of full adaptation framework designed to enhance the capabilities of SAM. We term this adaption technique as full adaption, since it adapts the neural network globally. In contrast to partial adaption of Med-SA, GMed-SA focuses more on adapter collaboration and therefore perform global adjustments to improve SAM's performance with less effort, thus it modifies the output embedding of SAM and aligns it to the medical domain characteristics faster and better. The Adapter Modules can be described as follows:

GMed-SA is built based on MLP-A, and it consists of a down-projection linear transformation layer to reduce dimension, a ReLU activation, and an up-projection linear transformation to restore input size. In the encoder: It is placed by adding a residual path for each ViT block, and it serves to adapt and fine-tune the embedding processed by the ViT blocks, potentially acting as a bottleneck to improve parameter efficiency and adaptability. By updating GMed-SA parameters and freezing the majority of the pre-trained model at the same time, GMed-SA can achieve efficient adaptation without compromising performance.

### D. Global-Local Medical SAM adaptor

Due to the complementary characteristic of GMed-SA and Med-SA, we propose a global-local medical SAM adaptor (i.e., GMed-SA), which performs parameter-efficient fine-tuning using both full and partial adaption. The schematic diagram is depicted in figure 2. As seen, GMed-SA is based on Med-SA, and it additionally adds GMed-SA to each ViT block in image encoder. Following Med-SA [4] in training, we freeze SAM parameters, and we update the adapters simultaneously.

$$e_{adapt} = g(e_{input}) + l(e_{input}), \quad (1)$$

where $e_{input}$ represents the input embedding of a ViT block; $e_{adapt}$ represents the adapted embedding by GL-Med-SA; $g(e_{input})$ represents the output of adapters of GMed-SA; and $l(e_{input})$ represents the corresponding output embedding of Med-SA in the ViT block.

## IV. EXPERIMENTAL RESULTS AND ANALYSIS

In this section, we first describe the experimental setup, evaluation metrics, training strategy, and implementation details. We then compare the performance of GL-Med-SA with other SOTA methods on challenging 2D melanoma dataset. Experimental results are shown and analyzed in terms of various evaluation metrics including Dice similarity coefficient (Dice) and intersection of union (IoU).

### A. Experimental Datasets and Metrics

We conducted experiments using the publicly accessible International Skin Imaging Collaboration (ISIC) dataset. Following Med-SA, the segmentation performance was evaluated using Dice and IoU. The input resolution is set to 1024×1024.

### B. Prompt Strategy for Interactive Segmentation

Our approach to interactive segmentation involves utilizing click prompts during the training phase. Following Med-SA, the click prompt generation revolves around the concept of using positive clicks to denote foreground areas one times or three times. That is, a hybrid strategy that randomly and iteratively samples foreground regions to simulate interactive clicks is utilized. Specifically, we start with random sampling to initialize the prompts, followed by an iterative approach that adds a few more clicks. This iterative method mimics real-user interaction by placing new clicks in the foreground areas where the model's predictions, based on previous clicks, have been incorrect.

### C. Implementation Details

For this research, we adopt Med-SA pipeline according to its official GitHub repository and add our adapters to the image encoder (i.e., a pre-trained ViT-H [9]). For two dimensional medical images, we maintained default configurations of Med-SA. All the involved datasets were subjected to a training regimen of 35 epochs for GL-Med-SA. The reason why we selected fewer epochs compared to that of Med-SA is that our model appeared to converge more rapidly in the same default setup of Med-SA. Note that, the prompt configuration comprises a single random positive point (referred to as "1-point") prompt and three random positive point (referred to as "3-point") prompt.

All experimental procedures were carried out using PyTorch deep learning framework and executed on a NVIDIA A800 graphic processing unit (GPU) under windows operation system. We stuck to the default settings in Med-SA for the comparison methods in experiments, as we use it as the base of our method.

TABLE I. THE COMPARISON OF GL-MED-SA WITH MED-SA AND SOTA SEGMENTATION METHODS ON MELANOMA

| Model | Param (M) | | Dice | IoU |
|---|---|---|---|---|
| | Frozen | trainable | | |
| SAM [1] 1-points | 636 | 0 | 81.6 | 70.4 |
| SAM [1] 3-points[a] | 636 | 0 | 85.8 | 77.5 |
| MedSAM [10] 1-point | 636 | 636 | 86.8 | 77.5 |
| MedSAM [10] 3-points | 636 | 636 | 87.5 | 78.6 |
| Med-SA [4] 1-point | 636 | 13 | 92.6 | 84.1 |
| Med-SA [4] 3-point | 636 | 13 | 93.4 | 84.7 |
| GL-Med-SA 1-point | 636 | 20 | 92.9 | 84.5 |
| GL-Med-SA 3-point | 636 | 20 | **93.8** | **85.0** |
| ResUNet [11] | 17 | 17 | 87.1 | 78.2 |
| BEAL [12] | 25 | 25 | 86.6 | 78.0 |
| TransBTS [13] | 39 | 39 | 88.1 | 80.6 |
| EnsemDiff [14] | 27 | 27 | 88.2 | 80.7 |
| MTSeg [15] | 19 | 19 | 87.5 | 79.7 |
| UltraUNet [16] | 19 | 19 | 89.0 | 81.8 |
| FAT-Net [17] | 75 | 75 | 90.7 | 83.9 |
| BAT [18] | 88 | 88 | 91.2 | 84.3 |
| SegDiff [19] | 32 | 32 | 87.3 | 79.4 |
| nnUNet [20] | 16 | 16 | 90.8 | 83.6 |
| TransUNet [21] | 96 | 96 | 89.4 | 82.2 |
| UNetr [22] | 104 | 104 | 89.7 | 82.8 |
| Swin-UNetr [23] | 138 | 138 | 90.2 | 83.1 |

a. "3 points" means the method randomly and iteratively samples a point prompt for three times.

## D. Melanoma Segmentation Performance Comparison

We compared several SOTA interactive segmentation methods with 1-point GL-Med-SA, where only a point was randomly selected as the point prompt, and 3-point GL-Med-SA, where only a point was randomly selected as the point prompt. The compared methods include the ones specifically designed for melanoma [17, 18] and several transformer based ones [11-16, 19-23]. As seen in Table 1, Both Dice and IoU performance indicate that our method achieves performance gain over the base model and other compared ones with same number of point prompts, demonstrating the superiority of our method. This is mainly due to the combination of global parameter efficient fine-tuning of GMSA and the strong transfer ability of Med-SA. We have also visualized good and failure cases in figure 3 to show more details.

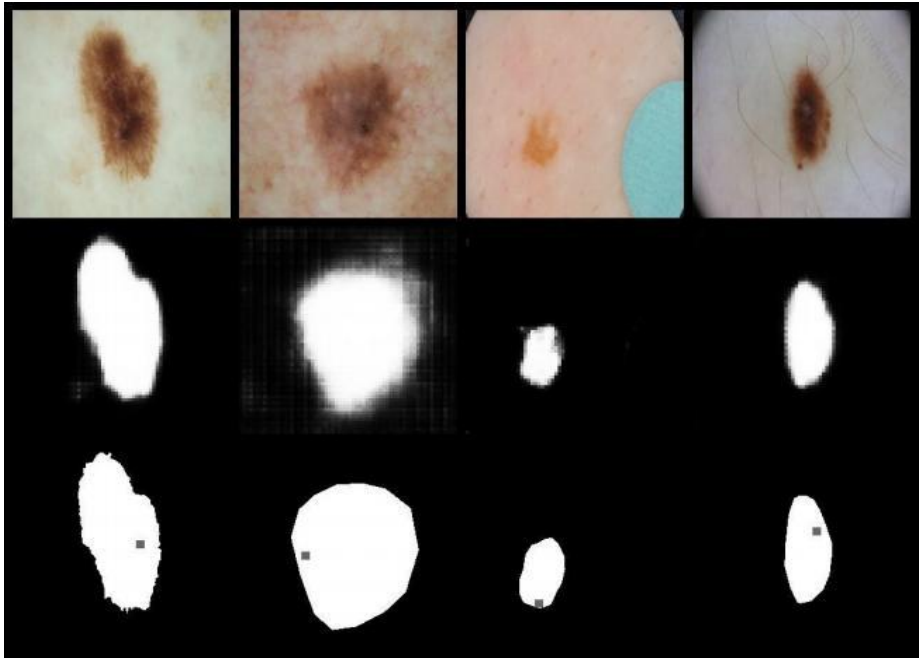

Figure 3. Visual performance of GLMed-SA. From left to right, the first, third and forth columns illustrate successful segmentation results of GL-Med-SA on melanoma dataset, and the third column shows a failure segmentation case. The gray blocks indicate the "1-point" prompts.

## V. DISCUSSION

By comparing GL-Med-SA with other SOTA methods on a challenging 2D melanoma dataset, we find that the performance of GL-Med-SA is comparable or even better than existing approaches, in terms of both Dice and IoU evaluation metrics. This may owe to the following reasons brought by GMed-SA:

Parameter Eficiency: GMed-SA requires only a small fraction of the parameters to be updated, significantly reducing computational cost and memory usage compared to full fine-tuning.

Generalization: The global embedding adjustments in GMed-SA enable the model to better generalize to diverse medical image modalities and segmentation tasks, enhancing its applicability in real-world scenarios.

Ease of Implementation: GMed-SA is relatively straightforward to implement and can be easily plugged into SAM-based frameworks, making it accessible for researchers and practitioners.

Moreover, GL-Med-SA is compatible to other visual language models, it presents a promising approach for advancing medical image segmentation, and it may unlock new possibilities in clinical applications.

## VI. CONCLUSION

GL-Med-SA presents a promising approach for advancing medical image segmentation and unlocking new possibilities in clinical applications. This is realized by By leveraging the power of partial adaption and full adaption. The characteristics of GL-Med-SA enable it can be applied to many potential applications, such as cell segmentation in histopathology images, facilitating quantitative analysis and disease characterization. Future research could explore the following directions to further enhance GMed-SA, including Investigating different PEFT techniques and Adapter designs to optimize performance and efficiency, integrating domain-specific knowledge and prior information into GMed-SA to improve segmentation accuracy in specific medical applications, and extending GMed-SA to handle multi-modal medical images, combining information from different imaging modalities to achieve more comprehensive segmentation results.

## ACKNOWLEDGMENT

We acknowledge the support of Liaoning Provincial Doctoral Research Startup Fund Program (project number 2024-BS-258).